# TeochewBench: A Human-Reviewed Benchmark for Teochew Hanzi Translation

Jianan Wu
Peking University
jnwu25@stu.pku.edu.cn

## Abstract

Teochew has a substantial speaker community and exhibits distinctive lexical, syntactic, and pragmatic features, yet textual resources for evaluating modern large language models remain limited. We present TeochewBench, a human-reviewed benchmark comprising 300 Teochew Hanzi expressions for evaluating translation from Teochew Hanzi into Mandarin Chinese and English. The dataset covers five categories: basic vocabulary; everyday sentences; Teochew-specific expressions; tone, politeness, and context; and idiomatic, ambiguous, and culturally specific expressions. A primary Teochew-speaking reviewer examined all entries individually and revised them as needed, while two additional Teochew speakers verified selected items.

Our main evaluation covers 11 official general-purpose post-trained models on the reviewed dataset in both translation directions, yielding 6,600 predictions. Two official base checkpoints provide a further 1,200 predictions for supplementary diagnostics, bringing the total to 13 models and 7,800 predictions. We additionally include a Hanzi-copy control, which returns the source input unchanged, to assess how shared Hanzi affect automatic scores for translation into Mandarin Chinese. Qwen3.5-27B achieved the highest overall chrF-style score among the evaluated checkpoints, at 60.63, followed by Qwen2.5-72B-Instruct at 56.61, Gemma-3-27B-IT at 56.36, and GLM-4-32B-0414 at 55.82. Across the 11 main-evaluation models, the mean chrF-style score decreased from 69.25 for low-specificity items to 27.52 for high-specificity items. High-specificity expressions received lower scores and exhibited smaller cross-model differences, suggesting that they constitute a shared low-scoring region across the model families evaluated here. The Hanzi-copy control further indicates that surface overlap in low-specificity items can substantially affect automatic scores for translation into Mandarin Chinese.

## 1 Introduction

Large language models have advanced rapidly on translation and language-understanding tasks in high-resource languages. Evaluation resources for regional languages and language varieties, however, are often constrained by limited data, variation in writing conventions and pronunciation, and a scarcity of reference outputs. Teochew is used extensively in everyday communication and exhibits a rich range of linguistic practices. Although Teochew Hanzi expressions share many surface forms with Mandarin Chinese, they also contain regionally distinctive vocabulary, constructions, sentence-final particles, idioms, and cultural meanings. Because the two varieties share a writing system without fully sharing meaning and usage, simple string-based metrics may capture both translation performance and surface overlap between the source and reference.

Existing resources such as Teochew-Wild focus on speech-processing tasks, including automatic speech recognition and text-to-speech synthesis (Pan et al., 2025). In contrast, structured and human-reviewed resources for evaluating large language models on translation from written Teochew into Mandarin Chinese and English remain limited. TeochewBench is designed to evaluate large language models on translation from Teochew Hanzi into Mandarin Chinese and English.

This work makes the following contributions:

1. We construct a 300-item evaluation set for translating Teochew Hanzi into Mandarin Chinese and English. The set covers five categories of linguistic content and includes one or more Mandarin Chinese and English references for each item.

2. We establish a documented human-review and revision workflow. A primary Teochew-speaking reviewer reviewed all 300 items, including the Teochew Hanzi expressions and their Mandarin Chinese and English references, while two additional Teochew speakers verified selected items.

3. We evaluate 11 official general-purpose post-trained models from five families under a common protocol across two translation directions, producing 6,600 predictions. Supplementary evaluations of two official base checkpoints contribute another 1,200 predictions.

4. We use textual-specificity stratification and a Hanzi-copy control to examine how shared Hanzi and Teochew-specific textual features relate to automatic evaluation scores.

## 2 Related Work

### 2.1 Teochew Resources

Teochew-Wild contains 18.9 hours of in-the-wild Teochew speech with orthographic transcriptions and Teochew romanization annotations, and supports research on automatic speech recognition and text-to-speech synthesis (Pan et al., 2025). Publicly accessible lexicographic resources such as PUJDICT provide information about Teochew vocabulary, pronunciation, and regional pronunciation variation (PUJDICT contributors, n.d.). These resources support speech processing and lexicographic lookup, but serve a different purpose from text translation evaluation sets for generative models. The latter require clearly defined translation directions, natural reference translations, and a consistent evaluation protocol. Because written Teochew shares many Hanzi with Mandarin Chinese, evaluation should also account for automatic-score gains attributable to copying the source at the surface level.

### 2.2 Evaluation of Dialects and Language Varieties

Prior work has evaluated language variation through controlled linguistic features rather than aggregate scores alone. VALUE constructs an African American Vernacular English (AAVE) variant of GLUE using lexical and morphosyntactic transformation rules, while Multi-VALUE extends this rule-based framework to 50 English varieties and 189 linguistic features (Ziems et al., 2022, 2023). For Sinitic machine translation, Yu et al. (2024) developed a manually translated FLORES+ benchmark from English into Wu Chinese, documenting a workflow for constructing an evaluation set centered on a specific Sinitic variety. Zheng and Bloem (2026) evaluate bidirectional translation for Fuzhounese using automatic metrics and human judgments. TeochewBench likewise reports stratified results rather than only aggregate scores: it stratifies items by textual specificity and examines the association between textual specificity and content category. The textual-specificity rubric counts independent Teochew-specific lexical, grammatical, pragmatic, or semantic features relative to the Mandarin Chinese reference; its highest tier also covers idioms, allusions, fixed cultural expressions, and non-literal meanings. The labels characterize textual distinctiveness relative to the Mandarin Chinese reference, rather than the correctness or naturalness of the Teochew expression or its translations.

### 2.3 Automatic Translation Metrics and Surface Overlap

Following prior work on character-level machine translation evaluation, including chrF (Popović, 2015), we report a locally implemented chrF-style character n-gram score, multiset character F1, and normalized exact match. None of these metrics requires word segmentation. However, for translation from Teochew Hanzi into Mandarin Chinese, the source and reference share many Hanzi, so returning the source unchanged can receive a substantial automatic score. We therefore include an explicit Hanzi-copy control for the Mandarin Chinese direction and report its scores separately across textual-specificity tiers. We use the automatic metrics to compare the relative performance of the evaluated checkpoints.

# 3 The TeochewBench Dataset

## 3.1 Scope and Tasks

TeochewBench evaluates text translation from Teochew Hanzi into Mandarin Chinese and English. Each item contains a Teochew Hanzi expression, one or more Mandarin Chinese reference translations, one or more English reference translations, a content-category label, and a textual-specificity label. Multiple English references may be provided to cover literal and meaning-based renderings, as well as natural translations appropriate to different contexts.

## 3.2 Candidate Construction

We initially compiled candidate items with reference to publicly accessible Teochew dictionaries and other language resources, including PUJDICT, using automated tools to assist with organization. The candidate items subsequently underwent human review. All Teochew Hanzi expressions, Mandarin Chinese references, and English references in the evaluation set used in this study reflect their human-reviewed versions. Automated tools were used to organize candidate materials. Final linguistic judgments and dataset content were determined through human review.

## 3.3 Human Review

A native speaker of Teochew served as the primary reviewer. The primary reviewer is from the Chaoshan region and has used Teochew regularly in family and everyday communication since childhood. The primary reviewer examined all 300 candidate items individually and revised them as needed. The review covered the naturalness and idiomaticity of the Teochew Hanzi expressions, the semantic alignment of the Mandarin Chinese and English reference translations, and the textual-specificity assignments.

Two additional native speakers of Teochew verified selected items. Their verification focused on Teochew-specific usages, slang, and cultural expressions, while also including items from other textual-specificity tiers. They primarily checked the Teochew expressions and their corresponding meanings in Mandarin Chinese. When disagreements arose, the primary reviewer considered the additional speakers' feedback and determined the final version. All reviewers agreed to participate in the study, and the manuscript does not disclose personally identifying information.

## 3.4 Category Composition

| **Category** | **Count** |
|---|---|
| Basic Vocabulary | 60 |
| Everyday Sentences | 100 |
| Teochew-Specific Expressions | 60 |
| Tone, Politeness, and Context | 50 |
| Idioms, Ambiguity, and Cultural Expressions | 30 |
| **Total** | **300** |

## 3.5 Teochew Textual Specificity

Textual specificity characterizes the extent to which a Teochew Hanzi expression exhibits Teochew-specific textual features relative to its Mandarin Chinese reference:

• `low`: contains no independent Teochew-specific textual feature and is nearly identical to the Mandarin Chinese reference at the character level;

• `ordinary`: contains one independent Teochew-specific lexical, grammatical, pragmatic, or semantic feature;

• `medium`: contains two such independent features;

• `high`: contains at least three such independent features, or involves slang, idioms or sayings, allusions, fixed cultural expressions, or non-literal meanings.

| Tier | Count | Percentage |
|---|---|---|
| `low` | 82 | 27.3% |
| `ordinary` | 146 | 48.7% |
| `medium` | 39 | 13.0% |
| `high` | 33 | 11.0% |

The labels characterize textual distinctiveness rather than the correctness or naturalness of either the Teochew expression or its reference translations. The primary reviewer assigned or confirmed a textual-specificity label for every item using this rubric.

# 4 Experimental Setup

## 4.1 Models

The main evaluation comprises 11 official general-purpose post-trained checkpoints from five families: Qwen, GLM, Gemma, Seed-OSS, and MiniCPM. From the locally available model pool, we select original-developer releases spanning multiple parameter scales; secondary task-specific fine-tunes are outside this comparison. Each checkpoint is evaluated on all 300 items in both translation directions, yielding 600 predictions per checkpoint and 6,600 main-evaluation predictions. Seven of the 11 main checkpoints belong to the Qwen family.

We additionally evaluate two official base checkpoints, Qwen3-4B-Base and Qwen3.5-4B-Base, for supplementary diagnostics reported in Appendix B. They contribute 1,200 predictions, giving 13 reported checkpoints and 7,800 model predictions in total. The base checkpoints are not included in the main model rankings, textual-specificity aggregates, cross-model dispersion statistics, or bootstrap comparisons in Sections 5 and 6. The 300 Hanzi-copy outputs are counted separately from model predictions.

## 4.2 Generation and Reproducibility

Across checkpoints, we used the same direction-specific task instructions, rendered through each model's chat template. We used greedy decoding with `temperature = 0`, `top_p = 1`, `top_k = 0`, and a maximum generation length of 128 tokens. The prompt instructed each model to return only the translation in the requested target language, without explanations, alternatives, labels, or quotation marks. Checkpoint-specific deployment settings were recorded for each run. Appendix A provides the task instructions, output-processing rules, and supplementary inference details.

### 4.3 Metrics and the Hanzi-Copy Control

We report three automatic metrics: a locally implemented chrF-style character n-gram score, multiset character F1, and normalized exact match. Before scoring, text is Unicode NFKC-normalized, case-folded, and stripped of non-alphanumeric characters. The chrF-style metric takes the arithmetic mean of available order-level character n-gram F-scores over orders 1-6 with $\beta = 2$ and is reported on a 0-100 scale. An order is excluded only when neither the prediction nor the reference contains an n-gram of that order; if only one side contains such n-grams, that order receives zero. Multiset character F1 is computed from character-count overlap and is reported on a 0-1 scale. Before final scoring, we separated pre-existing alternative Mandarin translations and editorial usage or spelling notes in 34 items, including three items with corresponding English reference cleanup. This reference-only revision left source texts and stored model predictions unchanged. For items with multiple references, each metric uses the highest score across the available references. Aggregate chrF-style and multiset character F1 scores are averaged over predictions, while normalized exact match is reported as an accuracy.

To estimate the contribution of shared Hanzi to automatic scores, the Hanzi-copy control returns the Teochew Hanzi input unchanged as the Mandarin Chinese output. This control is defined only for translation from Teochew Hanzi into Mandarin Chinese; it produces no English output and serves as a diagnostic control rather than a translation system.

We compute 95% confidence intervals using 2,000 bootstrap replicates, resampling source items within each compared tier and keeping the two target-language observations for each source together. Because most items have only one English reference, string-based metrics may under-reward valid paraphrases and synonymous translations.

## 5 Results

### 5.1 Overall Results

| Model | Mandarin Chinese chrF-style | English chrF-style | Overall chrF-style |
|---|---|---|---|
| Qwen3.5-27B | 60.21 | 61.06 | 60.63 |
| Qwen2.5-72B-Instruct | 55.44 | 57.78 | 56.61 |
| Gemma-3-27B-IT | 55.01 | 57.70 | 56.36 |
| GLM-4-32B-0414 | 55.92 | 55.73 | 55.82 |
| Qwen2.5-14B-Instruct | 54.24 | 53.65 | 53.94 |
| Qwen3.5-9B | 50.17 | 55.52 | 52.84 |
| Qwen3.5-4B | 50.70 | 53.02 | 51.86 |
| Qwen3-8B | 48.69 | 49.73 | 49.21 |
| Qwen2.5-7B-Instruct | 44.07 | 49.54 | 46.80 |
| Seed-OSS-36B-Instruct | 30.95 | 54.05 | 42.50 |
| MiniCPM4-0.5B | 26.27 | 25.08 | 25.67 |

Qwen3.5-27B achieved the highest overall chrF-style score among the evaluated checkpoints, at 60.63, followed by Qwen2.5-72B-Instruct at 56.61. Gemma-3-27B-IT and GLM-4-32B-0414 ranked third and fourth, at 56.36 and 55.82, respectively. Both achieved higher overall scores than Qwen2.5-14B-Instruct, indicating that relatively high automatic scores in this evaluation were not confined to Qwen checkpoints.

Across this heterogeneous checkpoint set, the results do not exhibit a simple monotonic relationship between model size and automatic scores. Differences in pretraining, post-training, and instruction-following behavior may all contribute to the observed results. Because several adjacent checkpoints have similar scores, we do not interpret small differences in automatic metrics as definitive evidence of capability differences.

The supplementary base-model results are reported separately in Appendix B and do not enter this ranking.

## 5.2 Textual Specificity

The following statistics use only the 11 main-evaluation checkpoints.

| Textual-specificity tier | Items | Mean chrF-style | Mean cross-model item SD | SD of model means | Mean pairwise model gap |
|---|---|---|---|---|---|
| `low` | 82 | 69.25 | 21.40 | 10.72 | 10.66 |
| `ordinary` | 146 | 47.56 | 23.01 | 9.97 | 11.28 |
| `medium` | 39 | 39.26 | 19.08 | 9.10 | 10.58 |
| `high` | 33 | 27.52 | 12.64 | 4.52 | 5.47 |

Mean chrF-style averages scores across the 11 main checkpoints and both target languages. For each item-target observation, cross-model item SD is the standard deviation of chrF-style scores across these checkpoints; the table reports its mean within each tier. All reported standard deviations are population standard deviations over the evaluated checkpoints. SD of model means is the standard deviation of the 11 checkpoint-level mean scores within a tier. Mean pairwise model gap is the mean absolute difference between those means over all 55 unordered checkpoint pairs.

Nine of the 11 main checkpoints exhibited a strictly decreasing mean chrF-style pattern, `low > ordinary > medium > high`, and the `high` tier received the lowest mean score for 10 of the 11 checkpoints. Cross-model dispersion was smallest in the `high` tier, but did not decrease monotonically across all four tiers. The `high`-minus-`ordinary` difference in mean cross-model item SD was -10.37 (95% CI [-13.45, -7.11]). These results suggest that high-specificity items constitute a shared low-scoring region for the main-evaluation checkpoints, rather than necessarily providing stronger discrimination among them.

## 5.3 Copying Effects

| Textual-specificity tier | Hanzi-copy EM | Hanzi-copy multiset character F1 | Hanzi-copy chrF-style |
|---|---|---|---|
| `low` | 0.524 | 0.870 | 67.69 |
| `ordinary` | 0.034 | 0.490 | 21.86 |
| `medium` | 0.000 | 0.443 | 11.23 |
| `high` | 0.000 | 0.339 | 8.16 |

Item-level Hanzi-copy chrF-style scores correlated negatively with the ordinal textual-specificity tiers (Spearman's $\rho = -0.634$). This association suggests that higher textual specificity generally corresponds to lower source-reference surface overlap. The Hanzi-copy chrF-style score of 67.69 for low-specificity items shows that surface character overlap can substantially affect automatic scores when translating from Teochew Hanzi into Mandarin Chinese. The Hanzi-copy control is defined only for the Mandarin Chinese direction and should not be compared directly with model averages pooled across Mandarin Chinese and English.

Textual specificity and content category had Cramér's V = 0.292 in this dataset. Tier-level differences may therefore partly reflect category composition and should not be attributed to textual specificity alone.

# 6 Discussion

## 6.1 Main Findings

The experiments yield two main findings. First, low-specificity items exhibit substantial surface overlap between the Teochew Hanzi source and the Mandarin Chinese reference, allowing the unchanged-source control to obtain relatively high automatic scores. Evaluations in the Mandarin Chinese direction should therefore report the Hanzi-copy control alongside model results to avoid interpreting character overlap as translation performance.

Second, automatic scores generally decrease as textual specificity increases across the 11 main-evaluation checkpoints. High-specificity items also exhibit the smallest cross-model dispersion, although dispersion does not decrease monotonically across all tiers. The high-specificity tier receives the lowest mean score for 10 of the 11 checkpoints, including GLM-4-32B-0414 and Gemma-3-27B-IT, suggesting a shared low-scoring region across multiple model families. However, because this tier currently contains only 33 items, the finding should be validated on larger and more balanced datasets.

## 6.2 Future Work

Future work will involve additional Teochew speakers from a broader range of regional backgrounds and will incorporate independent review and adjudication of disagreements. We will also conduct a more systematic blinded human evaluation using a larger, stratified sample of model outputs to examine how automatic metrics correspond to human judgments of translation quality. Dataset expansion will prioritize expressions with distinctive Teochew lexical, grammatical, pragmatic, and cultural features while improving balance across content categories and textual-specificity tiers. Items with weak Teochew-specific features, redundancy, or limited discriminative value will be revised or replaced based on subsequent review and experimental evidence, with all changes documented in the dataset's version history.

# 7 Limitations and Ethics Statement

The current evaluation set contains 300 short text items and cannot fully represent the regional, generational, orthographic, and pragmatic diversity of Teochew. A primary Teochew-speaking reviewer conducted a structured review of all 300 items, while two additional Teochew speakers verified selected items. Because no second reviewer conducted an independent review of the full dataset, we do not report inter-annotator agreement. The reported experiments evaluate only translation from Teochew Hanzi into Mandarin Chinese and English; spoken Teochew and Peng'im romanization are outside their scope.

We rely primarily on automatic metrics to evaluate model outputs. When translating from Teochew Hanzi into Mandarin Chinese, character-overlap metrics may reward copying from the source. String-based metrics in both translation directions may also penalize acceptable paraphrases and non-literal translations. Moreover, 240 of the 300 items currently have only one English reference translation. Textual specificity is associated with content category, and the high-specificity tier contains only 33 items. The stratified results therefore require validation on a larger and more balanced dataset.

The main evaluation covers five model families, but seven of its 11 checkpoints belong to the Qwen family, so coverage across families remains uneven. The two supplementary base checkpoints are both from Qwen and are not pooled with the main models. Although GLM and Gemma provide additional cross-family comparisons, future work should include a broader range of Chinese-focused and multilingual model families. More systematic human evaluation is also needed to determine whether small differences in automatic scores correspond to meaningful differences in translation quality.

Teochew varies across regions and communities, and TeochewBench should not be interpreted as defining a single authoritative or standardized form of Teochew. The textual-specificity labels describe differences relative to Mandarin Chinese references and should not be interpreted as judgments about the correctness or legitimacy of particular Teochew forms.

## 8 Conclusion

TeochewBench provides a shared evaluation framework for translating Teochew Hanzi into Mandarin Chinese and English. Across the 11 main-evaluation checkpoints, automatic scores generally decreased as textual specificity increased, with especially low scores on expressions containing multiple Teochew-specific textual features, culturally grounded meanings, or non-literal language. Gemma-3-27B-IT and GLM-4-32B-0414 ranked among the four highest-scoring checkpoints by overall chrF-style score, indicating that relatively high automatic scores in this evaluation were not confined to the Qwen family. The Hanzi-copy control further shows that aggregate automatic scores in the Mandarin Chinese direction can be substantially influenced by character overlap, particularly for low-specificity items. Future work will expand the dataset and review coverage and use more systematic human evaluation to examine how closely automatic metrics correspond to human judgments of translation quality.

## References

Pan, Linrong, Chenglong Jiang, Gaoze Hou, and Ying Gao. 2025. "Teochew-Wild: The First In-the-Wild Teochew Dataset with Orthographic Annotations." *arXiv preprint arXiv:2505.05056*. https://arxiv.org/abs/2505.05056

Popović, Maja. 2015. "chrF: Character n-gram F-score for Automatic MT Evaluation." In *Proceedings of the Tenth Workshop on Statistical Machine Translation*, 392-395. Association for Computational Linguistics. https://doi.org/10.18653/v1/W15-3049

PUJDICT contributors. n.d. "PUJDICT: Péh-Uē-Jī Dictionary of the Teochew-Swatow Dialect." Accessed August 28, 2026. https://github.com/pujdict/pujdict

Yu, Hongjian, Yiming Shi, Zherui Zhou, and Christopher Haberland. 2024. "Machine Translation Evaluation Benchmark for Wu Chinese: Workflow and Analysis." In *Proceedings of the Ninth Conference on Machine Translation*, 600-605. Association for Computational Linguistics. https://doi.org/10.18653/v1/2024.wmt-1.47

Zheng, Sue, and Jelke Bloem. 2026. "Benchmarking Multilingual LLM Translation Accuracy for Fuzhounese." In *Proceedings of the SIGUL 2026 Joint Workshop with ELE, EURALI, and DCLRL: Towards Inclusivity and Equality: Language Resources and Technologies for Under-Resourced and Endangered Languages*, 198-209. ELRA Language Resources Association. https://doi.org/10.63317/4mm9bs8yy4ie

Ziems, Caleb, Jiaao Chen, Camille Harris, Jessica Anderson, and Diyi Yang. 2022. "VALUE: Understanding Dialect Disparity in NLU." In *Proceedings of the 60th Annual Meeting of the Association for Computational Linguistics (Volume 1: Long Papers)*, 3701-3720. Association for Computational Linguistics. https://doi.org/10.18653/v1/2022.acl-long.258

Ziems, Caleb, William Held, Jingfeng Yang, Jwala Dhamala, Rahul Gupta, and Diyi Yang. 2023. "Multi-VALUE: A Framework for Cross-Dialectal English NLP." In *Proceedings of the 61st Annual Meeting of the Association for Computational Linguistics (Volume 1: Long Papers)*, 744-768. Association for Computational Linguistics. https://doi.org/10.18653/v1/2023.acl-long.44

# Appendix A. Prompting and Inference Details

## A.1 Prompt Construction

For every item and target language, we used the same logical two-message structure.

### System message

```
You are a careful translation engine. Return only the translation, with no explanation,
alternatives, labels, or quotation marks.
```

### User message for Mandarin Chinese

```
Translate the following Teochew text written in Chinese characters into natural Mandarin Chinese.
ID: {item_id}
Text: {teochew_hanzi}
Translation:
```

### User message for English

```
Translate the following Teochew text written in Chinese characters into natural English.
ID: {item_id}
Text: {teochew_hanzi}
Translation:
```

The only item-specific dataset fields supplied to the model were the item identifier and the Teochew Hanzi source. The target language was specified by the fixed task instruction. Reference translations, review notes, category labels, and textual-specificity labels remained on the scoring side and were not supplied to the models.

The messages were rendered using each checkpoint's tokenizer-specific chat template with `add_generation_prompt=True`. All reported runs used a model chat template, and the fully rendered prompt was retained with each prediction. The task wording and item-field structure were held fixed across checkpoints, although model-specific control tokens and template scaffolding differed across model families.

For the Qwen3 and Qwen3.5 checkpoints, as well as Qwen2.5-72B-Instruct and Seed-OSS-36B-Instruct, the template renderer was called with `enable_thinking=False` and `thinking_budget=0`; each template used the arguments it recognized. The remaining checkpoints were rendered without these additional reasoning-control arguments.

## A.2 Output Processing

Both the raw generation and the processed prediction were retained. Processing removed leading and trailing whitespace, handled closing `</seed:think>` and `</think>` tags by retaining the text following the final closing tag, removed at most one leading translation label, and removed supported enclosing quotation marks when the first and last characters matched. No semantic rewriting, target-language correction, or manual correction was applied. Malformed and empty processed outputs remained in the evaluation, with empty outputs receiving zero automatic scores.

## A.3 Inference and Scoring Details

All 13 reported checkpoints, comprising the 11 main models and two supplementary base checkpoints, were evaluated locally using vLLM 0.22.1 and Transformers 5.11.0, with locally stored weights loaded in offline mode. The shared inference configuration was as follows:

| Setting | Value |
|---|---|
| Precision | bfloat16 |
| Temperature | 0 |
| Top-p | 1.0 |
| Top-k | 0 |
| Maximum new tokens | 128 |
| Maximum model length | 1,024 |
| Random seed | 0 |
| Prompt format | Checkpoint-specific chat template |
| Predictions per checkpoint | 600 |

Each checkpoint produced one prediction for each combination of 300 items and two target languages. Tensor parallelism and memory allocation were adjusted to accommodate model size and available hardware; these deployment settings were not treated as experimental variables.

The metric implementation is described in Section 4.3. For Mandarin Chinese, 270 items had one scoring reference, 28 had two, and two had three. For English, 240 items had one reference, 56 had two, and four had three. For items with multiple references, normalized exact match, multiset character F1, and chrF-style independently selected the highest-scoring reference; the maximizing reference could therefore differ across metrics.

The reported percentile bootstrap confidence intervals used 2,000 replicates with seed 20260825 and were computed using only the 11 main-evaluation checkpoints. For the textual-specificity comparison, source items were resampled with replacement within each compared tier; both target-language observations were kept together for each sampled source item.

# Appendix B. Supplementary Official Base-Model Diagnostics

## B.1 Scope and Results

We report Qwen3-4B-Base and Qwen3.5-4B-Base as supplementary official base checkpoints. Each produced 600 predictions using the same source items, direction-specific instructions, generation limits, output-processing rules, reference translations, and scoring metrics as the main evaluation. Their 1,200 predictions are included in the reported total of 7,800, but excluded from the 6,600 main-evaluation predictions and all cross-model statistics in the main text.

| Model | Mandarin Chinese chrF-style | English chrF-style | Overall chrF-style |
|---|---|---|---|
| Qwen3-4B-Base | 2.83 | 3.73 | 3.28 |
| Qwen3.5-4B-Base | 45.84 | 50.13 | 47.98 |

Qwen3.5-4B-Base achieved an overall chrF-style score of 47.98, compared with 51.86 for the official post-trained Qwen3.5-4B checkpoint in the main evaluation, a descriptive difference of 3.88 points. This comparison is checkpoint-specific and does not isolate the effects of any individual post-training method.

## B.2 Interpretation

Qwen3-4B-Base was a clear outlier: frequent prompt echoes, repeated tags, and degenerate continuations indicate that its score reflects unreliable instruction following in addition to translation performance.

These diagnostics characterize the behavior of base checkpoints under the instruction-based evaluation protocol. Their scores should not be interpreted as pure measures of Teochew knowledge or pooled with the main post-trained model statistics. Qwen3-4B-Base has no same-size official post-trained Qwen3-4B counterpart in the reported model set, so it does not provide a matched-size post-training comparison here.